\pdfoutput=1

\documentclass[11pt]{article}

\usepackage[]{acl}

\usepackage{times}
\usepackage{latexsym}
\usepackage{graphicx}
\usepackage[nolist]{acronym}

\usepackage[T1]{fontenc}

\usepackage[utf8]{inputenc}

\usepackage{microtype}

\newcommand*{\TitleFont}{%
      \usefont{\encodingdefault}{\rmdefault}{b}{n}%
      \fontsize{14}{14}%
      \selectfont}

\title{\TitleFont A Decade of Knowledge Graphs in Natural Language Processing: A Survey}

\author{Phillip Schneider$^1$, Tim Schopf$^1$, Juraj Vladika$^1$, Mikhail Galkin$^2$, \\ {\bf Elena Simperl$^3$ and Florian Matthes$^1$} \\
         $^1$Technical University of Munich, Department of Computer Science, Germany \\
         $^2$Mila Quebec AI Institute \& McGill University, School of Computer Science, Canada \\ 
         $^3$King’s College London, Department of Informatics, United Kingdom \\
         \texttt{\{phillip.schneider, tim.schopf, juraj.vladika, matthes\}@tum.de} \\
         \texttt{mikhail.galkin@mila.quebec} \\
         \texttt{elena.simperl@kcl.ac.uk}}

\begin{acronym}
    \acro{rq}[RQ]{research question}
    \acroplural{rq}[RQs]{research questions}
    \acro{nlp}[NLP]{natural language processing}
    \acro{plm}[PLM]{pretrained language model}
    \acroplural{plm}[PLMs]{pretrained language models}
    \acro{ai}[AI]{artificial intelligence}
    \acro{qa}[QA]{Question answering}
    \acro{kbqa}[KBQA]{question answering over knowledge bases}
    \acro{nlg}[NLG]{Natural language generation}
    \acro{kg}[KG]{knowledge graph}
    \acroplural{kg}[KGs]{knowledge graphs}
\end{acronym}

\begin{document}

\maketitle
\begin{abstract}
In pace with developments in the research field of artificial intelligence, knowledge graphs (KGs) have attracted a surge of interest from both academia and industry. As a representation of semantic relations between entities, KGs have proven to be particularly relevant for natural language processing (NLP), experiencing a rapid spread and wide adoption within recent years. 
Given the increasing amount of research work in this area, several KG-related approaches have been surveyed in the NLP research community. However, a comprehensive study that categorizes established topics and reviews the maturity of individual research streams remains absent to this day. 
Contributing to closing this gap, we systematically analyzed 507 papers from the literature on KGs in NLP. Our survey encompasses a multifaceted review of tasks, research types, and contributions. As a result, we present a structured overview of the research landscape, provide a taxonomy of tasks, summarize our findings, and highlight directions for future work.

\end{abstract}

\begin{figure*}[ht!]
  \includegraphics[width=\textwidth]{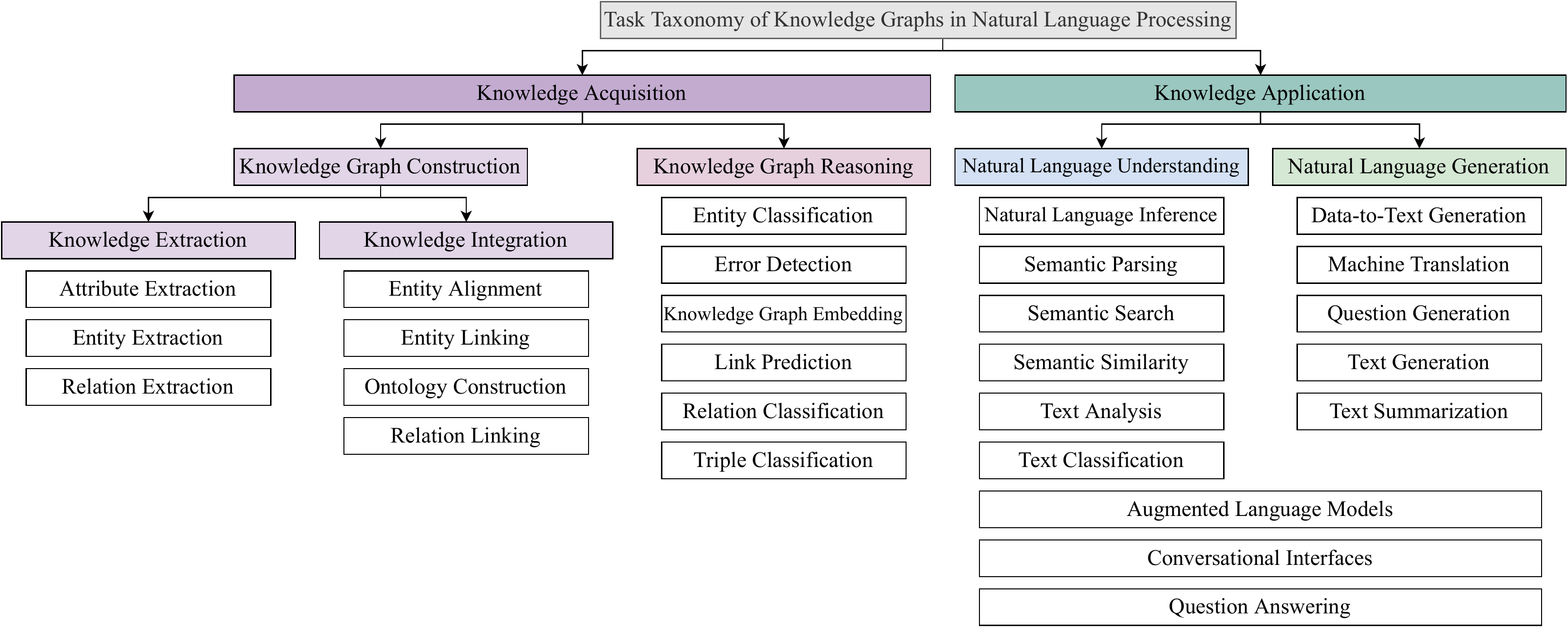}
  \caption{Taxonomy of tasks in the literature on KGs in NLP.}
  \label{fig:taxonomy}
\end{figure*}

\section{Introduction}
\label{sec:introduction}
Knowledge acquisition and application are inherent to natural language. Humans use language as a means of communicating facts, arguing about decisions, or questioning beliefs. Therefore, it is not surprising that computational linguists started already in the 1950s and 60s to work out ideas on how to represent knowledge as relations between concepts in semantic networks \cite{Richens1956PreprogrammingFM,Quillian_Ross,COLLINS1969240}.

More recently, \acp{kg} have emerged as an approach for semantically representing knowledge about real-world entities in a machine-readable format. They originated from research on semantic networks, domain-specific ontologies, as well as linked data, and are thus not an entirely new concept \cite{10.1145/3397512}. Despite their growing popularity, there is still no general understanding of what exactly a \ac{kg} is or for what tasks it is applicable. Although prior work has already attempted to define \acp{kg} \cite{10.1007/978-3-642-41335-3_34,Ehrlinger2016TowardsAD,paulheim2017knowledge,10.3233/SW-170275}, the term is not yet used uniformly by researchers. Most studies implicitly adopt a broad definition of \acp{kg}, where they are understood as \textit{"a graph of data intended to accumulate and convey knowledge of the real world, whose nodes represent entities of interest and whose edges represent relations between these entities"} \cite{Hogan_2022}. 

\acp{kg} have attracted a lot of research attention in both academia and industry since the introduction of Google’s KG in 2012 \cite{singhal_2012}. Particularly in \ac{nlp} research, the adoption of \acp{kg} has become increasingly popular over the past 5 years, and this trend seems to be accelerating. The underlying paradigm is that the combination of structured and unstructured knowledge can benefit all kinds of \ac{nlp} tasks. For instance, structured knowledge from \acp{kg} can be injected into that of the contextual knowledge found in language models, which improves the performance in downstream tasks \cite{DBLP:journals/corr/abs-2101-12294}. Furthermore, with the growing importance of \acp{kg}, there are also expanding efforts to construct new \acp{kg} from unstructured texts.

Ten years after Google coined the term knowledge graph in 2012, a plethora of novel approaches has been proposed by scholars. Therefore, it is important to assemble insights, consolidate existing results, and provide a structured overview. However, to our knowledge, there are no studies that offer an overview of the whole research landscape of \acp{kg} in the \ac{nlp} field. Contributing to closing this gap, we performed a comprehensive survey to analyze all research performed in this area by classifying established topics, identifying trends, and outlining areas for future research. Our three main contributions are as follows:
\begin{enumerate}
    \item We systematically extract information from 507 included papers and report insights about tasks, research types, and contributions. 
    \item We provide a taxonomy of tasks in the literature on \acp{kg} in \ac{nlp} shown in Figure \ref{fig:taxonomy}.
    \item We assess the maturity of individual research streams, identify trends, and highlight directions for future work.
\end{enumerate}

Our survey sheds light on the evolution and current research progress regarding \acp{kg} in \ac{nlp}. Although we cannot achieve complete coverage of all relevant papers on this topic, we aim at providing a representative overview that can help both \ac{nlp} scholars and practitioners by offering a starting point in the literature. Moreover, our multifaceted analysis can guide the research community in closing existing gaps and finding novel ways how to combine \acp{kg} with \ac{nlp}.

\section{Related Work}
\label{sec:related-work}
Related literature that includes both \acp{kg} and \ac{nlp} seems to be relatively scarce. Most survey papers focus either only on \acp{kg} or only on \ac{nlp}. In their broad introduction to \acp{kg}, \citet{Hogan_2022} point out that existing surveys on \acp{kg} tend to revolve around specific aspects of \acp{kg}, most commonly their construction and embedding.

Such surveys with a KG focus usually bring up \ac{nlp} only in the context of employed \ac{nlp} methods, like information extraction, being used to populate and refine graphs \citep{Nickel_2016}. Other surveys on \acp{kg} mention some downstream applications of \acp{kg} for \ac{nlp} tasks, such as for constructing augmented language models, \ac{kbqa}, or recommender systems \citep{ji2021survey}.

As noted previously, related work that includes both \acp{kg} and \ac{nlp} strictly focus on a specific application or task. For example, \citet{safavi-koutra-2021-relational} provide an overview on applying relational world knowledge from \acp{kg} to augment large contextual language models. Other surveys on specific applications include \ac{kg} reasoning \citep{chen2019}, biomedical \acp{kg} \citep{NICHOLSON20201414}, and the task of \ac{kbqa} \citep{DBLP:journals/corr/abs-2007-13069}.

The survey on graphs in \ac{nlp} by \citet{nastase2015survey} covers only smaller graphs such as dependency graphs and dialogue trees. Even though it does not include \acp{kg}, the survey concludes that graphs are a powerful representation formalism and how \ac{nlp} tasks can benefit from harnessing the potential of data presented in graph structures.

To the best of our knowledge, this is the first survey covering a wide spectrum of techniques, methods as well as applications of \acp{kg} within the \ac{nlp} research field.

\section{Method}
To achieve our objective of providing a thorough overview of the research landscape, we conducted a systematic mapping study following the process defined by \citet{petersen2008systematic}. Its three main steps are explained in the next subsections.
\subsection{Research Questions}
The goal of our study is a multifaceted analysis of \acp{kg} in the field of \ac{nlp}, such as identifying and quantifying research topics, domains, and outcomes. These objectives are reflected in the \acp{rq} stated below. 

\textbf{RQ1}: What are the characteristics and trends of the research literature on \acp{kg} in \ac{nlp}?

\textbf{RQ2}: What are the different tasks mentioned in the existing research studies?

\textbf{RQ3}: What are the research types and main contributions of the studies?

\subsection{Search and Screening Procedure}
After specifying the \acp{rq}, we defined a set of related keywords for \acp{kg} and \ac{nlp} to be used for the database search of relevant studies. From initial test searches, we observed that including terms associated with \acp{kg} (e.g., “semantic network” or “ontology”) yielded too many irrelevant results. To restrict the research scope to the concept of \acp{kg}, we decided to use the following search string: 

\textit{("knowledge graph") AND ("NLP" OR "natural language processing" OR "computational linguistics")}. The search string was applied to title, abstract, and keywords. If a given paper had no keywords, we used index keywords from the database if they were available.

For our search of relevant publications, we queried six academic databases, as listed in Table \ref{tab:e-databases}. The ACL Anthology is a digital archive of prestigious conferences and journals in \ac{nlp}. ACM and IEEE provide access to publications of additional reputable venues in the broader computer science field. The remaining databases are commonly chosen in other related surveys to further increase the coverage of the respective field of interest.

In the first week of 2022, we applied our search string to the databases and restricted the time window to ten years from 2012 until 2021. Then, the exported files were merged, ensuring that each publication record was either a conference or a journal paper. We automatically identified and removed duplicate records as well. Through this, we obtained a dataset of 746 unique papers. Given this initial dataset, we further filtered down the truly relevant studies by screening for the following inclusion criteria: (1) peer-reviewed studies from conferences or journals, (2) studies with a clear focus on \acp{kg} in \ac{nlp}, (3) studies are written in English and full texts are electronically accessible. In reverse, this implies the publications that did not satisfy all three inclusion criteria were excluded from the dataset.

As part of the screening procedure, two of the authors read title, abstract, and keywords to determine if a paper matched the inclusion criteria. In ambiguous cases, the full text of the paper was examined. The two authors screened all papers and decided together on keeping or dropping records from the dataset. The final dataset included a total of 507 papers, as listed in Table \ref{tab:e-databases}. We make our annotated dataset available through a public GitHub repository.\footnote{\href{https://github.com/sebischair/KG-in-NLP-survey}{https://github.com/sebischair/KG-in-NLP-survey}}

\begin{table}
    \centering
    \begin{tabular}{lc}
    \hline
    \textbf{Academic Database} & \textbf{No. of Papers}\\
    \hline
    ACL Anthology & 164 \\
    ACM Digital Library & 26 \\
    IEEE Xplore & 76 \\
    ScienceDirect & 34 \\
    Scopus & 200 \\
    Web of Science & 7 \\
    \hline
    \textbf{Total} & \textbf{507} \\
    \hline
    \end{tabular}
    \caption{Overview of academic databases and number of included papers.}
    \label{tab:e-databases}
\end{table}

\subsection{Classification Scheme and Data Extraction}
According to our \acp{rq}, the included papers had to be categorized with respect to three facets: task, research type, and contribution. Established classification schemes from \citet{wieringa2006requirements} and \citet{shaw2003writing} were adapted for the research and contribution type as presented in Appendix \ref{sec:appendix-a}. For classifying tasks, we constructed a task taxonomy, following the iterative procedure suggested by \citet{ petersen2008systematic}, in which an initial classification scheme derived from keywords continuously evolves through adding, merging, or splitting categories during the classification process. Our task taxonomy is based on existing schemes from \citet{paulheim2017knowledge}, \citet{liu2020preliminary}, and \citet{ji2021survey}. Once the initial schemes were set up, all papers were sorted into the classes as part of the data extraction process. The 507 included studies were divided between two of the authors. In regular sessions, they discussed changes to the classification schemes or clarified uncertain labels. While each paper got assigned one label for the research type assigned, multiple labels were possible with regard to tasks and contributions. To assess the reliability of the inter-annotator agreement, the two authors independently classified a random sample of 50 papers. We calculated Cohen’s Kappa coefficient of these annotations for each facet \citep{cohen1960coefficient}. The annotations of the task, research, and contribution facets had coefficients of 0.73, 0.87, and 0.76, respectively. Cohen suggested interpreting Kappa values from 0.61 to 0.80 as substantial and from 0.81 to 1.00 as almost perfect agreement.

\section{Results}
In this chapter, we report the results of the data extraction process. It is arranged into subsections according to the formulated \acp{rq}.

\subsection{Characteristics of the Research Landscape (RQ1)}

\begin{figure}[ht]
  \includegraphics[width=0.99\columnwidth]{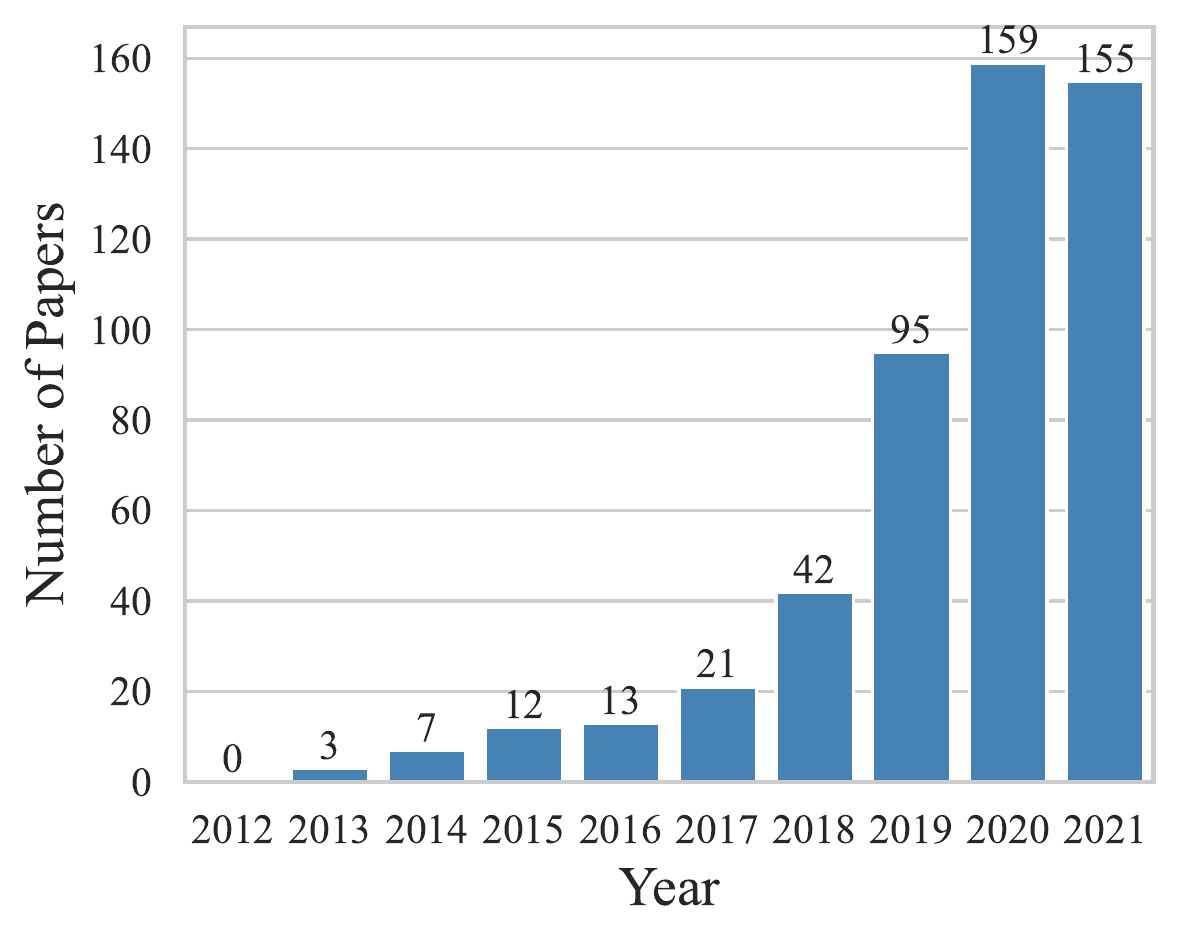}
  \caption{Distribution of number of papers from 2012 to 2021 (database export was performed in the first week of the year 2022).}
  \label{fig:paper-bar}
\end{figure}

In regard to the literature on \acp{kg} in \ac{nlp}, we started our analysis by looking at the number of studies as an indicator of research interest. The distribution of publications over the ten-year observation period is illustrated in Figure \ref{fig:paper-bar}. While the first publications appear in 2013, the annual publications grew slowly between 2013 and 2016. From 2017 onwards, the number of publications doubled almost every year. Because of the significant rise in research interest within these years, more than 90\% of all included publications originate from these five years. Even though the growth trend seems to stop in 2021, this is likely due to the data export which happened in the first week of 2022, leaving out many studies from 2021 that were enlisted in the databases later in 2022. Nonetheless, the trend in Figure \ref{fig:paper-bar} clearly indicates that \acp{kg} are receiving increasing attention from the \ac{nlp} research community. Considering the 507 included papers, the number of conference papers (402) was nearly four times as high as that of journal papers (105).

We also investigated institutional affiliations by country to determine what countries are most active in the field of \acp{kg} in \ac{nlp}. In total, we identified 44 countries contributing to the research literature. As part of the Appendix, we provide a world map with all countries in Figure \ref{fig:countries-geo} and a list of the top 20 countries by the number of affiliated papers in Table \ref{tab:countries-table}. China ranks first and holds a major proportion with 199 papers, accounting for 39\% of all publications. The United States and India come in second and third with 119 and 49 papers, respectively. Germany, the United Kingdom, and Italy follow in the ranking. All European countries had a combined total of 141 affiliated publications.

Another finding of the data extraction process concerns the diverse application areas of \acp{kg} in \ac{nlp}. We observed that the number of domains explored in the research literature grew rapidly in parallel with the annual count of papers. To reveal the great variety of areas, we list all 20 discovered domains and their subdomains in Table \ref{tab:subdomain-table} in the Appendix. In Figure \ref{fig:domains-bar}, the ten most frequent domains are displayed. It is striking that health is by far the most prominent domain. The latter appears more than twice as often as the scholarly domain, which ranks second. Other popular areas are engineering, business, social media, or law. In view of the domain diversity, it becomes evident that \acp{kg} are naturally applicable to many different contexts, as has been stated in prior work \cite{abu2021domain, ji2021survey, zou2020survey}.

\begin{figure}[h]
  \includegraphics[width=0.99\columnwidth]{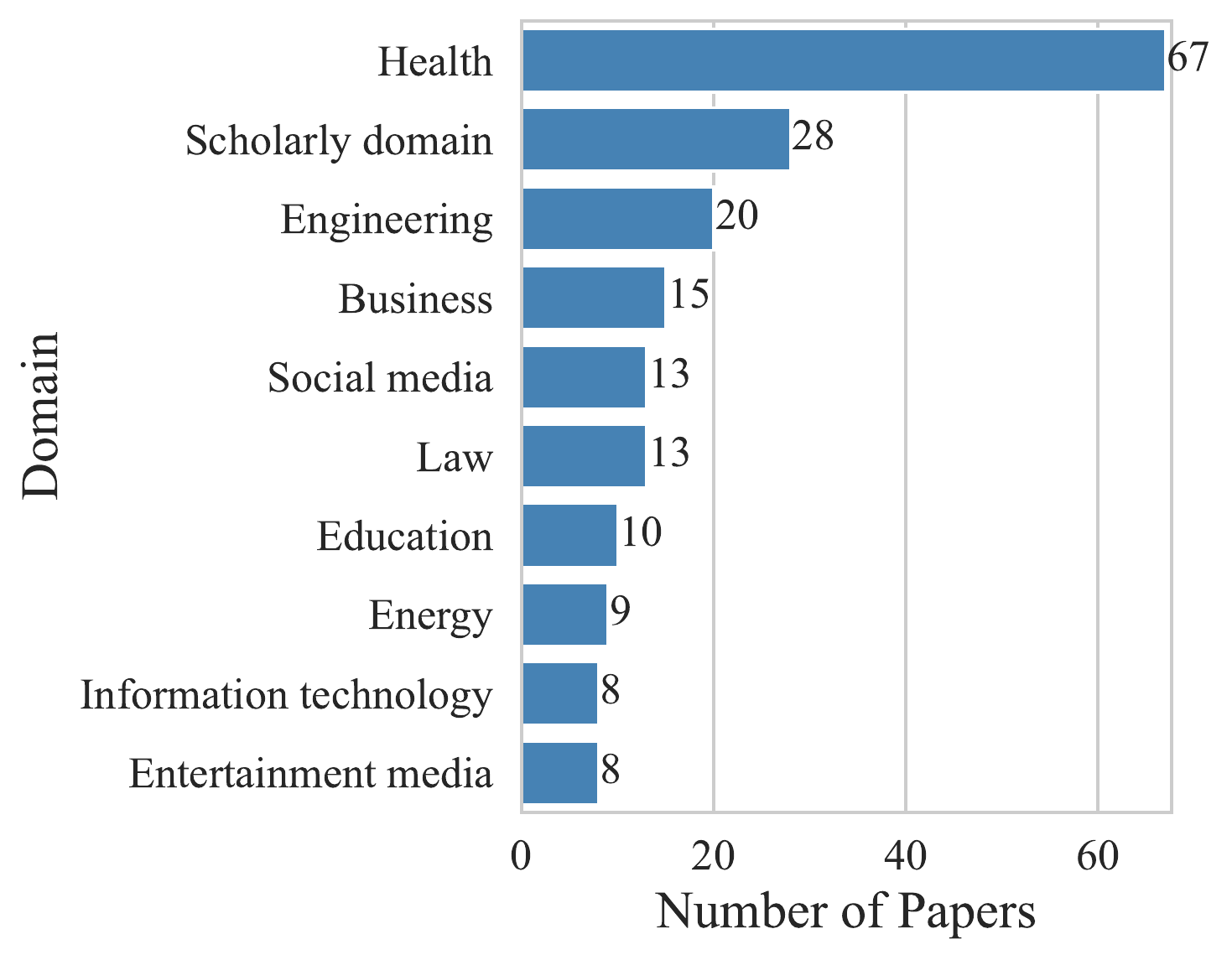}
  \caption{Number of papers by most popular application domains.}
  \label{fig:domains-bar}
\end{figure}

\begin{table*}[ht]
    \centering
    \begin{tabular}{lcp{7.8cm}}
    \hline
    \textbf{Task} & \textbf{No. of Papers} & \textbf{Representative Papers}\\
    \hline
    Relation extraction & 144 &
    \small
    \citet{peng-etal-2017-cross},
    \citet{wang-etal-2018-label}, \citet{zhang-etal-2019-long} \\
    Entity extraction & 143 &
    \small
    \citet{10.1016/j.websem.2015.12.004},
    \citet{luan-etal-2018-multi}, \citet{Wang2018InformationEA}\\
    Question answering & 103 & 
    \small
    \citet{bao-etal-2016-constraint}, \citet{10.5555/3504035.3504780}, \citet{feng-etal-2020-scalable} \\
    Semantic search & 91 & 
    \small
    \citet{10.5555/3298023.3298212}, \citet{10.1162/qss_a_00021}, 
    \citet{9357868} \\
    Augmented language models & 84 &
    \small
    \citet{zhang-etal-2019-ernie}, \citet{bosselut-etal-2019-comet}, \citet{Liu2020KBERTEL} \\
    Knowledge graph embedding & 61 &
    \small
    \citet{Shi_Weninger_2018}, \citet{ali2020benchmarking}, \citet{wang-etal-2021-kepler} \\
    Entity linking & 38 &
    \small
    \citet{kartsaklis-etal-2018-mapping}, \citet{moon-etal-2018-multimodal-named}, \citet{ijcai2018-556} \\
    Ontology construction & 32 & 
    \small
    \citet{gangemi2016framester}, \citet{foodkg}, \citet{10.1115/1.4046807} \\
    Conversational interfaces & 29 &
    \small
    \citet{10.5555/3304222.3304413} \citet{moon-etal-2019-opendialkg}, \citet{wu-etal-2019-proactive} \\
    Link prediction & 26 &
    \small
    \citet{lv-etal-2019-adapting}, \citet{sun-etal-2020-evaluation}, \citet{Wang_2021}\\
    \hline
    \end{tabular}
    \caption{Overview of most popular tasks in the literature on \acp{kg} in \ac{nlp}.}
    \label{tab:task-overview-table}
\end{table*}

\subsection{Tasks in the Research Literature (RQ2)}

Based on the tasks identified in the literature on \acp{kg} in \ac{nlp}, we developed the empirical taxonomy shown in Figure \ref{fig:taxonomy}. The two top-level categories consist of knowledge acquisition and knowledge application. Knowledge acquisition contains \ac{nlp} tasks to construct \acp{kg} from unstructured text (knowledge graph construction) or to conduct reasoning over already constructed \acp{kg} (knowledge graph reasoning). \ac{kg} construction tasks are further split into two subcategories: knowledge extraction, which is used to populate \acp{kg} with entities, relations, or attributes, and knowledge integration, which is used to update \acp{kg}. Knowledge application, being the second top-level concept, encompasses common \ac{nlp} tasks, which are enhanced through structured knowledge from \acp{kg}.

\begin{figure}[ht!]
  \includegraphics[width=0.99\columnwidth]{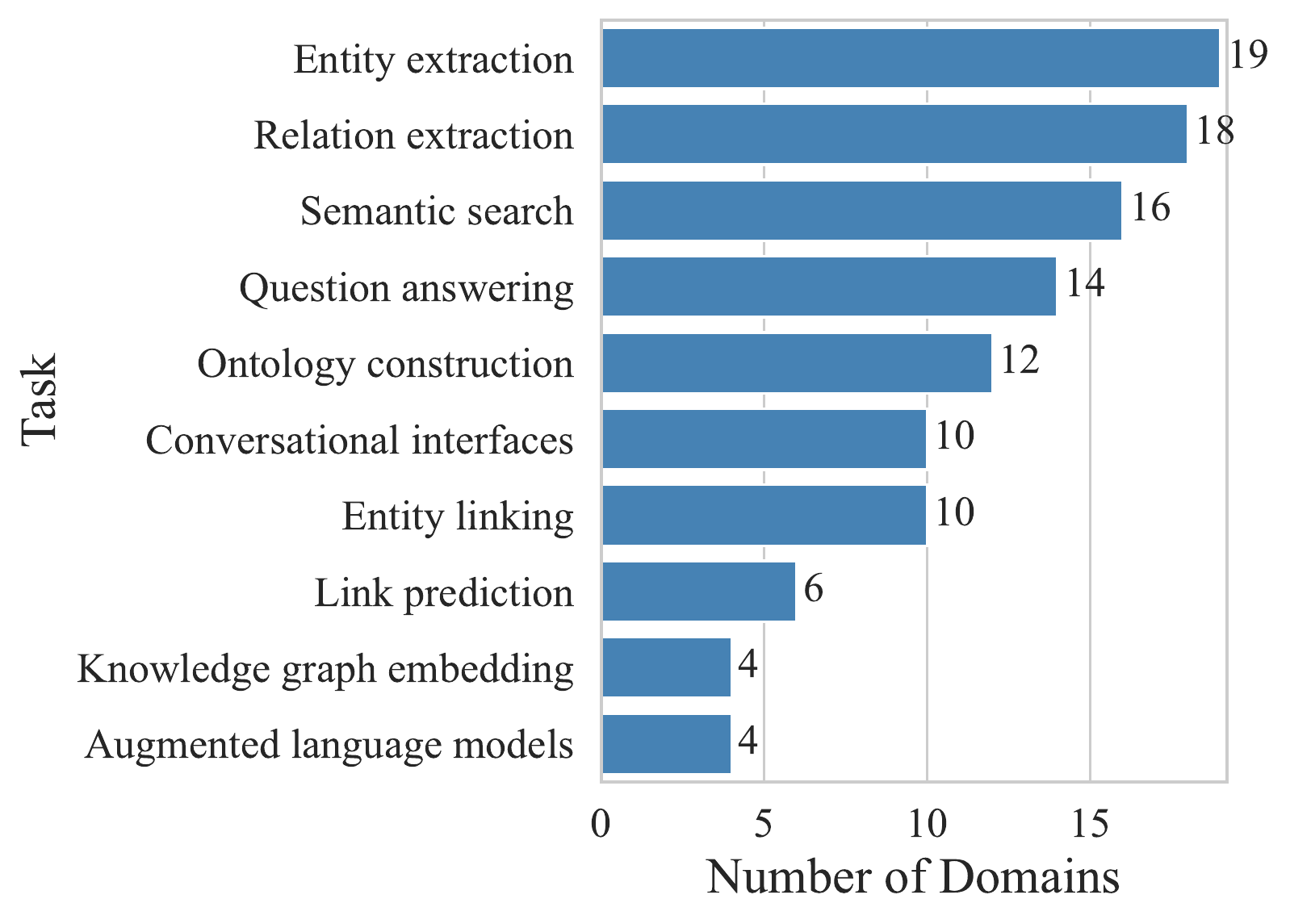}
  \caption{Overview of most popular tasks by number of application domains.}
  \label{fig:task-domains-bar}
\end{figure}

\begin{figure*}
  \includegraphics[width=\textwidth]{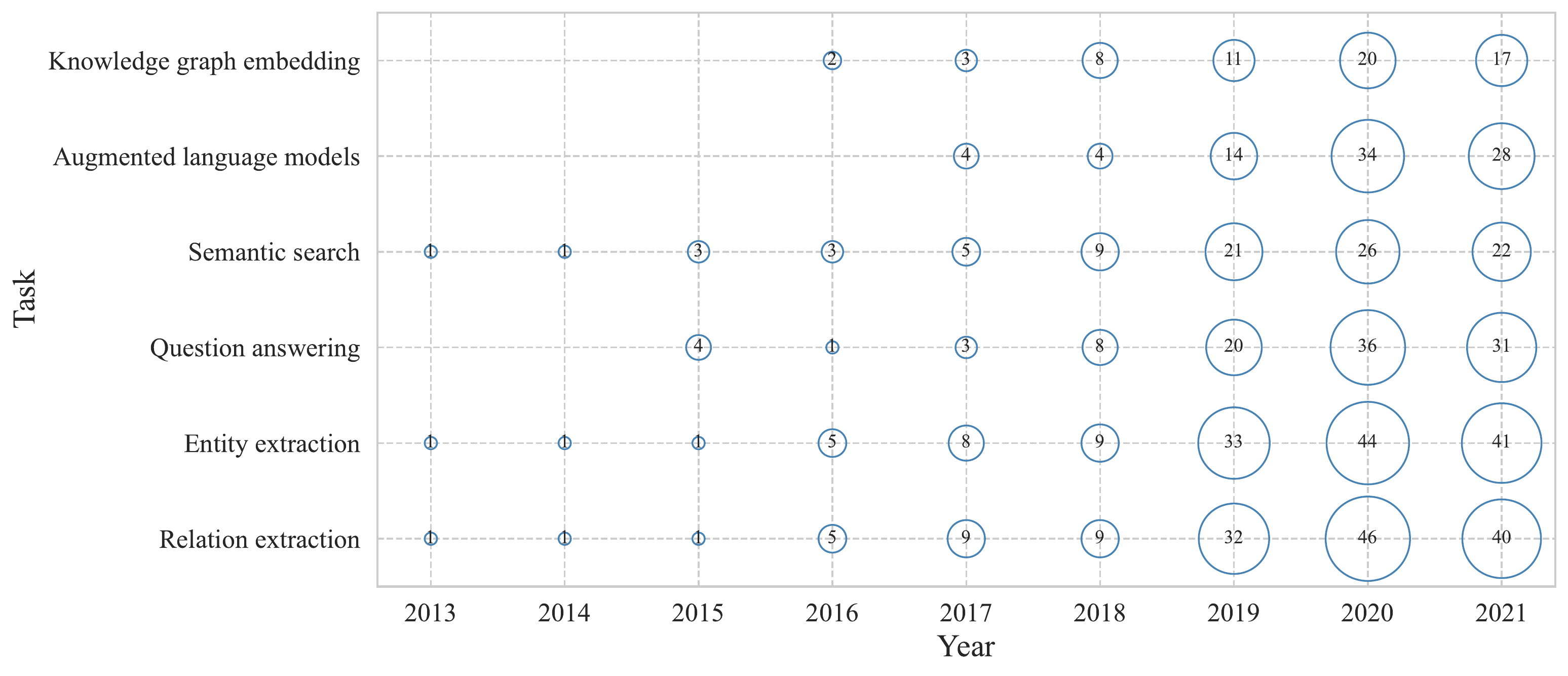}
  \caption{Distribution of number of papers by most popular tasks from 2013 to 2021.}
  \label{fig:task-bubble}
\end{figure*}

As might be expected, the frequency of occurrence in the literature for the tasks from our taxonomy varies greatly. While Table \ref{tab:task-overview-table} gives an overview of the most popular tasks, Figure \ref{fig:task-bubble} compares their popularity over time. Figure \ref{fig:task-domains-bar} displays the number of detected domains for the most prominent tasks. It shows that certain tasks are adopted to more domain-specific contexts than others.

\subsubsection{Knowledge Graph Construction}
The task of \textbf{entity extraction} is a starting point in constructing \acp{kg} and is used to derive real-world entities from unstructured text \citep{8999622}. Once the relevant entities are singled out, relationships and interactions between them are found with the task of \textbf{relation extraction} \citep{zhang-etal-2019-long}. A lot of papers use both entity extraction and relation extraction to construct new \acp{kg}, e.g., for news events \citep{10.1016/j.websem.2015.12.004} or scholarly research \citep{luan-etal-2018-multi}. 

\textbf{Entity linking} is a task of linking entities recognized in some text to already existing entities in \acp{kg} \citep{moon-etal-2018-multimodal-named, wu-etal-2020-scalable}. Since synonymous or similar entities often exist in different \acp{kg} or in different languages, \textbf{entity alignment} can be performed to reduce redundancy and repetition in future tasks \citep{gangemi2016framester, ijcai2018-556}. Coming up with the rules and schemes of \acp{kg}, i.e., their structure and format of knowledge presented in it, is done with the task of \textbf{ontology construction} \citep{foodkg}.

\subsubsection{Knowledge Graph Reasoning}
Once constructed, \acp{kg} contain structured world knowledge and can be used to infer new knowledge by reasoning over them. Thereby, the task of classifying entities is called \textbf{entity classification}, while \textbf{link prediction} is the task of inferring missing links between entities in existing \acp{kg} often performed via ranking entities as possible answers to queries \cite{Shi_Weninger_2018,bosselut-etal-2019-comet,wang-etal-2019-tackling,ali2020benchmarking}.

\textbf{Knowledge graph embedding} techniques are used to create dense vector representations of a graph so that they can then be used for downstream machine learning tasks. While this problem can be related solely to \acp{kg}, in our survey this label refers to approaches that jointly learn text and graph embeddings \citep{ijcai2018-556, wang-etal-2021-kepler}.

\subsubsection{Knowledge Application}
Existing \acp{kg} can be used in a multitude of popular \ac{nlp} tasks. Here we outline the most popular ones.

\textbf{\ac{qa}} was found to be the most common \ac{nlp} task using \acp{kg}. This task is typically divided into textual \ac{qa} and \acf{kbqa}. Textual \ac{qa} derives answers from unstructured documents while \ac{kbqa} does so from predefined knowledge bases \citep{DBLP:journals/corr/abs-2007-13069}. \ac{kbqa} is naturally tied to \acp{kg} while textual \ac{qa} can also be approached by using \acp{kg} as a source of common-sense knowledge when answering questions. As \citet{zhu2021} conclude, this approach is desired not only because it is helpful for generating answers, but also because it makes answers more interpretable.

\textbf{Semantic search} refers to "search with meaning", where the goal is not just to search for literal matches, but to understand the search intent and query context as well \citep{bast2016semantic}. This label denoted studies that use \acp{kg} for search, recommendations, and analytics. Examples are a big semantic network of everyday concepts called ConceptNet \citep{10.5555/3298023.3298212} and a \ac{kg} of scholarly communications and the relationships, among them the Microsoft Academic Graph \citep{10.1162/qss_a_00021}.

\textbf{Conversational interfaces} constitute another \ac{nlp} field that can benefit from world knowledge contained in \acp{kg}. \citet{10.5555/3304222.3304413} utilize the knowledge from \acp{kg} to generate responses of conversational agents that are more informative and appropriate in a given context. Knowledge-aware dialogue generation was also explored by \citet{moon-etal-2019-opendialkg}, \citet{wu-etal-2019-proactive}, \citet{liu-etal-2019-knowledge}.

\textbf{\ac{nlg}} is a subfield of \ac{nlp} and computational linguistics that is concerned with models which generate natural language output from scratch. \acp{kg} are used in this subfield for producing natural language text from \acp{kg} \citep{koncel-kedziorski-etal-2019-text}, generating question-answer pairs \citep{reddy-etal-2017-generating}, the multi-modal task of image captioning \citep{lu-etal-2018-entity}, or data augmentation in low-resource settings \citep{sharifirad-etal-2018-boosting}.

\textbf{Text analysis} combines various analytical \ac{nlp} techniques and methods that are applied to process and understand textual data. Exemplary tasks are sentiment detection \citep{kumar-etal-2018-knowledge}, topic modeling \citep{li2019integration}, or word sense disambiguation \citep{kumar-etal-2019-zero}.

\textbf{Augmented language models} are a combination of large \acp{plm} such as BERT \citep{devlin-etal-2019-bert} and GPT \citep{radford2018improving} with knowledge contained in \acp{kg}. Since \acp{plm} derive their knowledge from huge amounts of unstructured training data, a rising research trend is in combining them with structured knowledge. Knowledge from \acp{kg} can be infused into language models in their input, architecture, output, or some combination thereof \citep{DBLP:journals/corr/abs-2101-12294}. Some notable examples we outline are ERNIE \citep{zhang-etal-2019-ernie}, COMET \citep{bosselut-etal-2019-comet}, K-BERT \citep{Liu2020KBERTEL}, and KEPLER \citep{wang-etal-2021-kepler}.

\subsection{Research Types and Contributions (RQ3)}
\label{sec:RQ3}

Table \ref{tab:accents} shows the distribution of papers according to the different research and contribution types as defined in Table \ref{tab:research-type-scheme} and \ref{tab:contribution-type-scheme} in the Appendix. It shows that most papers conduct validation research, investigating new techniques or methods that have not yet been implemented in practice. A considerable number of papers, although significantly less, focus on solution proposals of approaches by demonstrating their advantages and applicability by a small example or argumentation. However, these papers usually lack a profound empirical evaluation. Secondary research accounts for only a small number of papers and is severely underrepresented in the research field of \acp{kg} in \acs{nlp}. As already mentioned in Section \ref{sec:introduction} and Section \ref{sec:related-work}, there is a notable lack of studies that summarize, compile, or synthesize existing research regarding \acp{kg} in \ac{nlp}. Moreover, evaluation research papers that implement and evaluate approaches in an industry context are equally scarce. Opinion papers are almost non-existent.

In terms of contribution types, techniques, methods, and tools are predominant. Resources and guidelines, as opposed to this, are rather underrepresented. This is in accordance with the distribution of research types, which indicates that mainly new methods and techniques are researched, but hardly any secondary research is conducted. Additionally, the research area of \acp{kg} in \ac{nlp} is lacking new resources such as text corpora, benchmarks, or constructed graphs.

\begin{table}[ht]
    \centering
    \begin{tabular}{lc}
    \hline
    \textbf{Research Type} & \textbf{No. of Papers}\\
    \hline
    Validation research & 338 \\
    Solution proposal & 149 \\ 
    Secondary research & 10 \\ 
    Evaluation research & 7 \\
    Opinion paper & 3 \\
    \hline
    \textbf{Contribution Type} & \textbf{No. of Papers}\\
    \hline
    Technique & 186 \\
    Method & 154 \\ 
    Tool & 139 \\ 
    Resource & 50 \\ 
    Guidelines & 24 \\ 
    \end{tabular}
    \caption{Number of papers by research type and contribution type.}
    \label{tab:accents}
\end{table}

\begin{figure}[ht]
  \includegraphics[width=0.99\columnwidth]{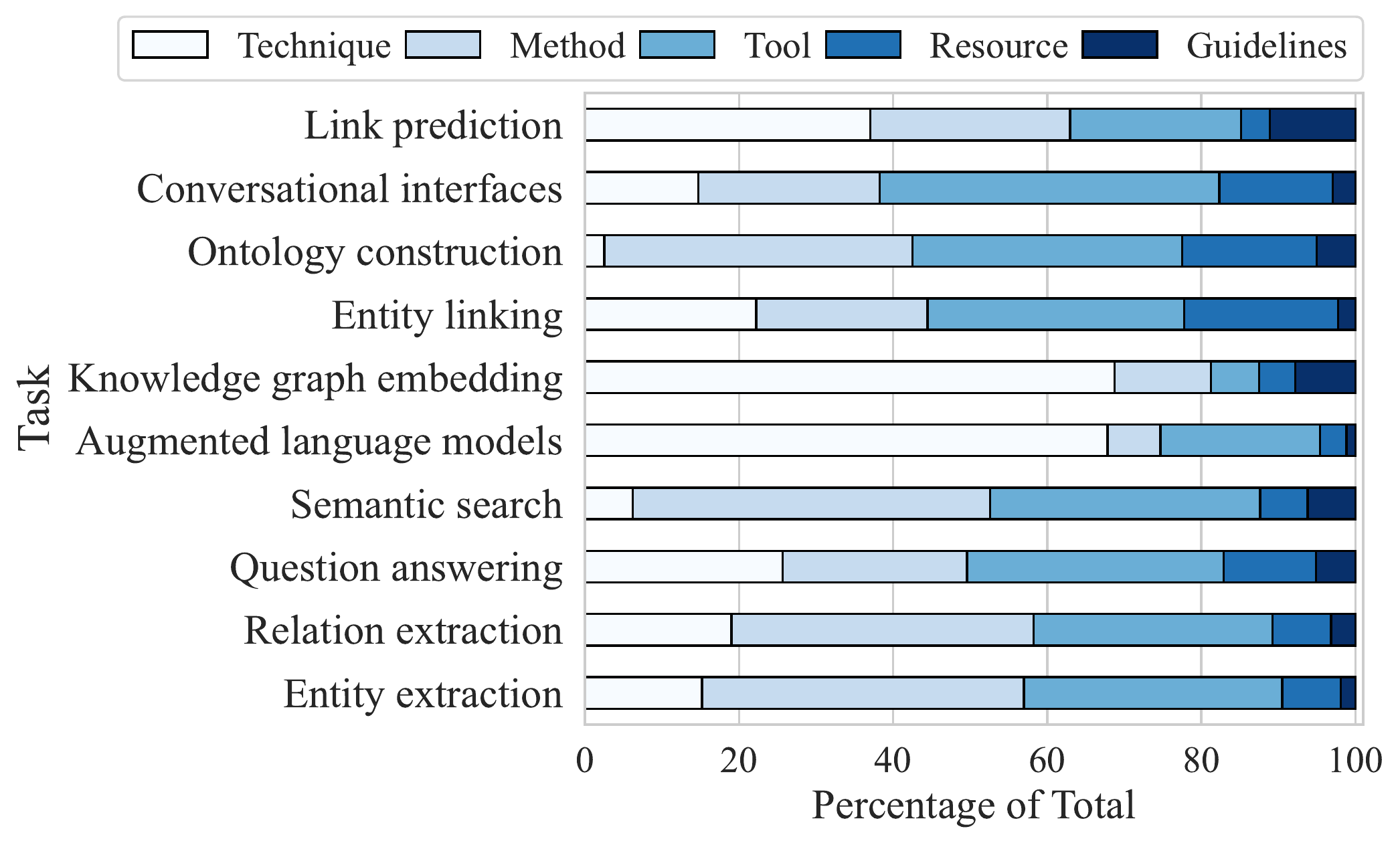}
  \caption{Percentage of contribution type by tasks.}
  \label{fig:stacked-tasks}
\end{figure}

Figure \ref{fig:stacked-tasks} depicts the different tasks of the analyzed studies and their relative share of contribution types. We can notice that entity extraction and relation extraction, which encompass the most works in line with Table \ref{tab:task-overview-table}, have a very balanced distribution of contribution types. These tasks, which build the foundation for \ac{kg} construction, have been researched for a long time and the number of studies in these areas is continually increasing, as can be seen in Figure \ref{fig:task-bubble}. Furthermore, a comparison of Figure \ref{fig:task-bubble} with Figure \ref{fig:stacked-tasks} shows that tasks, such as relation extraction or semantic search, which have existed for some time and continue to grow steadily have a rather balanced ratio of contribution types, too. This is an indication that these tasks are already reasonably mature, as some extensive preliminary work is required, for example, to use multiple techniques in a new method. 

Additionally, mature research areas already focus on industrialization, investigating how to use techniques in different domains and developing tools. Figure \ref{fig:task-domains-bar} strengthens the impression that tasks such as relation extraction or semantic search are already reasonably mature, as they are used in many different domains. In contrast, immature research areas still primarily focus on investigating new techniques and are used in a few domains only. For instance, the augmented language models and knowledge graph embedding tasks have mainly techniques as the contribution type and are not used in many different domains. Therefore, they can still be considered relatively immature. This may be a result of the fact that these tasks are still relatively young and less investigated. Figure \ref{fig:task-bubble} shows that the two tasks have only seen a sharp increase in studies from 2018 onwards and attracted a lot of interest since then. 

\section{Discussion}
The observations of our comprehensive survey reveal several insights. It is important to situate these findings with respect to related work and industry reports in the artificial intelligence (AI) field.

Since the first publications in 2013, researchers worldwide
have paid increasing attention to study \acp{kg} from a \ac{nlp} perspective, especially in the past five years. This observed growth in research interest is in line with the \ac{kg} survey of \citet{CHEN2021497}. We identified China and the United States as the most active countries shaping the research landscape, which is to be expected considering both countries regularly claim the top ranks in the popular "AI Index Report" from Stanford University \cite{DBLP:journals/corr/abs-2103-06312}. The report further highlights a soaring AI investment in the health domain. The latter was also the most dominant domain in our results (see Figure \ref{fig:domains-bar}). However, research in the health domain has to be considered critically, since these works compare poorly to other domains regarding reproducibility metrics, such as dataset and code accessibility \cite{10.1126/scitranslmed.abb1655}.

Table \ref{tab:accents} shows evidently that the research field of \acp{kg} in \ac{nlp} is lacking new resources such as text corpora, benchmarks, or \acp{kg}. This leads to the assumption that most works train and evaluate using the same limited available datasets and benchmarks. As a result, novel approaches are often optimized only for certain available benchmarks which may not hold up in practice. Furthermore, the lack of secondary research visible in Table \ref{tab:accents} reveals the need for more works that present an overview of the research field.

The frequency of tasks in our survey greatly varies, as reflected in Table \ref{tab:task-overview-table}. Studies concerning \ac{kg} construction account for the majority of all papers. Applied \ac{nlp} tasks such as \ac{qa} and semantic search also have a strong research community. The most emergent topics in recent years have been augmented language models, \ac{qa}, and \ac{kg} embedding. Some of the outlined tasks are still confined to the research community, while others have found practical application in many real-life contexts. From Figure \ref{fig:task-domains-bar} it is evident that the \ac{kg} construction tasks and semantic search over \acp{kg} are the most widely applied ones. Of the \ac{nlp} tasks, \ac{qa} and conversational interfaces have been adopted to many real-life domains, usually in the form of digital assistants. Tasks like \ac{kg} embedding and augmented language models are still only being researched and lack a widespread practical adoption in real-world scenarios. We anticipate that as the research areas of augmented language models and \ac{kg} embedding mature, more methods and tools will be investigated for these tasks.

\section{Limitations} 
Although we employed a rigorous study design and paid careful attention to executing each search and analysis step, our study is subject to limitations.

Given the restriction to one search string and six databases, there should be some relevant publications that we did not retrieve. This is the case for studies that did not mention our search terms in title, abstract, or keywords. To mitigate the risk of incompleteness, we chose common databases with a large number of publications in the examined research area. Further, we performed a preliminary search to optimize the completeness of results. Whenever possible, we replaced missing keywords with index keywords from the source database.

Moreover, the screening for relevant studies depends on the personal assessment of the researchers, which can bias the study selection. As a countermeasure, we defined selection criteria for the inclusion and exclusion of studies. During the study selection, two researchers assessed of selection criteria in parallel and discussed contradicting decisions until they reached a consensus to mitigate subjective bias.

The accuracy of the classification results constitutes another threat to the validity of our study. Data extraction bias may negatively affect the accuracy of the classification results. To mitigate this risk, the authors regularly discussed the used classification schemes and assigned labels to establish a common understanding of each class. In addition, we calculated Cohen's Kappa coefficient to quantify the reliability of the inter-annotator agreement.

\section{Conclusion}
Recent years have witnessed a rising prominence of \acp{kg} in \ac{nlp} research. Despite the rapidly growing body of literature, until now, no study has been published that summarizes the progress so far. To provide an overview of this maturing research area, we performed a multifaceted survey of tasks, research types, and contributions.

Our findings show that a large number of tasks concerning \acp{kg} in \ac{nlp} have been studied across various domains, including emerging topics like knowledge graph embedding or augmented language models. However, we observed a lack of secondary research and evaluations in practice, both of which are crucial to reflect the major scientific progress of the field as a whole. Our study lays the grounds for further research in this direction.

\section*{Acknowledgements}
This work has been supported by the German Federal Ministry of Education and Research (BMBF) Software Campus grant 01IS17049. The fourth author is partially supported by the Canada CIFAR AI Chair Program and Samsung AI grant (held at Mila).

\bibliography{anthology,custom}
\bibliographystyle{acl_natbib}

\appendix

\section{Supplementary Material}
\label{sec:appendix-a}
Table \ref{tab:research-type-scheme} shows the classification scheme for research types. \citet{wieringa2006requirements} introduced this scheme in order to categorize different types of research papers with differing approaches to what is being studied. Although the categories of evaluation research and validation research seem to be similar, there is a key difference. A paper is considered to be evaluation research only if the investigated problem is implemented and evaluated in practice. Papers labeled as validation research investigate properties of proposed solutions that have not been implemented in practice, while solution proposal papers introduce new solutions without a rigorous empirical validation.

Table \ref{tab:contribution-type-scheme} shows the classification scheme of contribution types employed in this study. It is based on the classification scheme of \citet{shaw2003writing} and adapted to the field of \acp{kg} in \ac{nlp}. Here, special attention needs to be paid to the distinction between method and technique. While a technique concentrates on solving a single specific task, a method involves a set of different techniques as well as procedures that must be executed in a systematic way to achieve a concrete objective.

Table \ref{tab:subdomain-table} contains an overview of the 20 domains we discovered in the literature on \acp{kg} in \ac{nlp}. For each domain, we identified a set of subdomains, which is listed as well.

\begin{figure*}[t]
  \includegraphics[width=16cm]{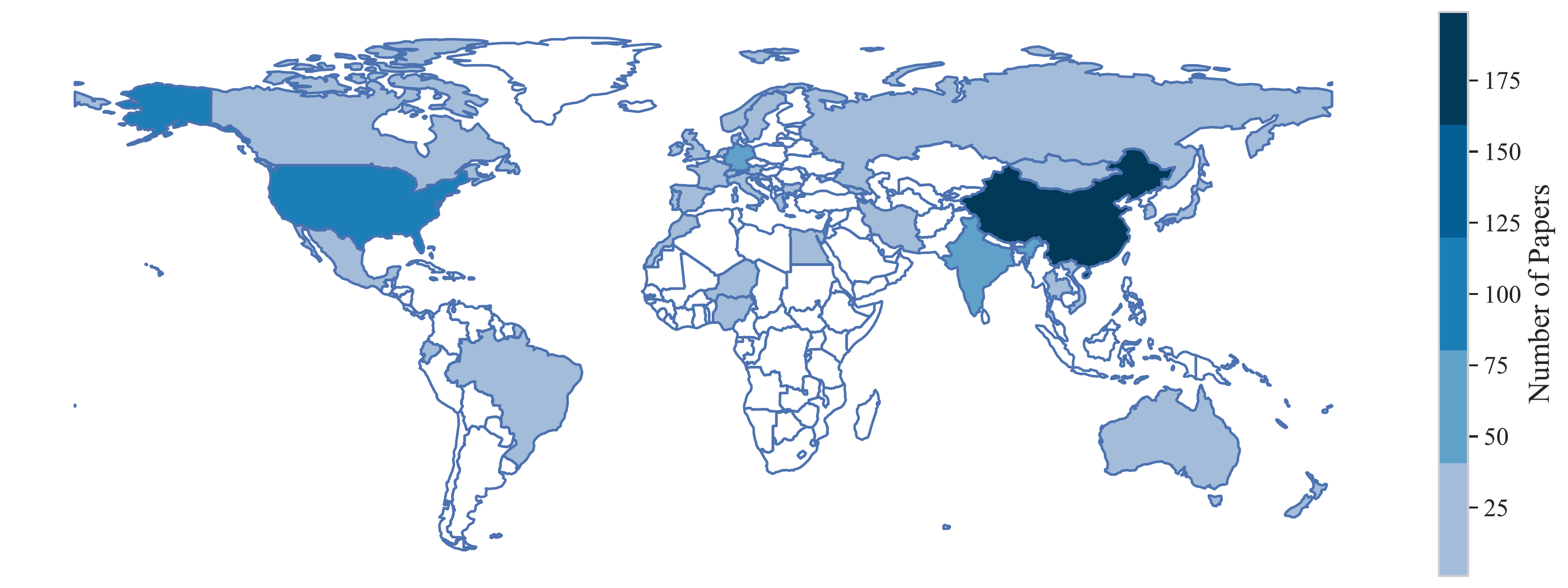}
  \caption{Global overview of number of papers by affiliated country.}
  \label{fig:countries-geo}
\end{figure*}

Table \ref{tab:countries-table} and the world map in Figure \ref{fig:countries-geo} give information about the number of papers by affiliated countries. While the table only shows the top 20 most active countries, the world map presents a global overview of all 44 countries contributing to the research literature.

\begin{table*}[t]
    \centering
    \begin{tabular}{lp{12cm}}
    \hline
    \textbf{Research Type} & \textbf{Description}\\
    \hline
    Evaluation research & The implementation of an existing technique or method is evaluated in practice within an industry context.\\
    Opinion paper & Report of the personal opinion of somebody on the suitability of a certain technique or method without relying on related work and research methods.\\
    Secondary research &  Analysis and synthesis of findings from multiple studies to systematically review a research field or gather evidence on a topic.\\
    Solution proposal & Proposal of novel solution or extension for a technique or method by demonstrating their advantages and applicability by a small example or argumentation.\\
    Validation research & Empirical investigation of characteristics from proposed techniques or methods that have not been implemented in practice yet.\\
    \hline
    \end{tabular}
    \caption{Classification scheme for research types adapted from \citet{wieringa2006requirements}.}
    \label{tab:research-type-scheme}
\end{table*}

\begin{table*}[t]
    \centering
    \begin{tabular}{lp{12cm}}
    \hline
    \textbf{Contribution Type} & \textbf{Description}\\
    \hline
    Guidelines &  List of advices or recommendations derived from the obtained research results.\\
    Method & A method contains a set of techniques and procedures  that need to be systematically executed to achieve a concrete goal.\\
    Resource &  A resource is a published data set that supports techniques, methods, or tools, e.g., text corpora, benchmarks, or knowledge graphs.\\
    Technique & A technique is the manner in which a concrete task within our task taxonomy is performed, often in the form of an algorithm or mathematical model.\\
    Tool & A tool is a documented implementation of a technique or method in the form of a software library, prototype, or full application system.\\
    \hline
    \end{tabular}
    \caption{Classification scheme for contribution types adapted from \citet{shaw2003writing}.}
    \label{tab:contribution-type-scheme}
\end{table*}

\begin{table*}[ht]
    \centering
    \begin{tabular}{lp{11cm}}
    \hline
    \textbf{Domain} & \textbf{Identified Subdomains}\\
    \hline
    Agriculture & Agricultural production, agricultural plant species\\
    Business & E-commerce, finance, human resources, product design, real estate\\
    Culture & Cultural heritage, ethnic minorities, film culture, museums, poetry \\
    Education & Curriculum design, digital library, e-learning, moral education  \\
    Energy & Oil and gas industry, power grid fault disposal, smart grid\\
    Engineering & Mechanical engineering, software engineering, electrical engineering \\
    Entertainment media & Computer games, media recommendation, movies, music, television \\
    Food & Dietary choices, recipe search \\
    Health & Biomedicine, traditional Chinese medicine, pharmacology, mental health \\
    History & Genealogy, historical events, retrieval of historical documents\\
    Information technology & App ecosystems, Internet of Things, technical support, cybersecurity \\
    Law & Law enforcement, patents, privacy policies, identity fraud detection\\
    Natural science & Mineralogy, oceanography, petroleum geology \\
    Scholarly domain & Bibliometrics, grant datasets, research collaborations, scientific corpora \\
    News & Fake news detection, journalism, news exploration \\
    Public sector & Government, military, poverty reduction, public safety organizations\\
    Social media & Insight extraction from posts, misinformation detection, opinion mining \\
    Social science & Open-source social science, social network analysis \\
    Sports & Basketball, football\\
    Tourism & Tourism question answering system, travel guide\\
    \hline
    \end{tabular}
    \caption{Overview of identified application domains and subdomains.}
    \label{tab:subdomain-table}
\end{table*}

\begin{table*}[b]
    \centering
    \begin{tabular}{clc}
    \hline
    \textbf{Rank} & \textbf{Country} & \textbf{No. of Affiliated Papers}\\
    \hline
    1 & China & 199 \\
    2 & United States & 119 \\
    3 & India & 49 \\
    4 & Germany & 47 \\
    5 & United Kingdom & 34 \\
    6 & Italy & 21 \\
    7 & Canada & 19 \\
    8 & Spain & 16 \\
    9 & France & 15 \\
    10 & Singapore & 14 \\
    11 & Australia & 13 \\
    12 & Hong Kong & 10 \\
    13 & Ireland & 9 \\
    14 & Netherlands & 8 \\
    15 & Japan & 8 \\
    16 & South Korea & 6 \\
    17 & Switzerland & 6 \\
    18 & Greece & 5 \\
    19 & Brazil & 5 \\
    20 & Portugal & 4 \\
    \hline
    \end{tabular}
    \caption{Overview of top 20 countries by number of affiliated papers.}
    \label{tab:countries-table}
\end{table*}

\end{document}